# Survey on Emotion Recognition through Posture Detection and the possibility of its application in Virtual Reality

**Leina Elansary** **leina.saad@cis.asu.edu.eg**
Master's student,Faculty of Computer and Information Sciences, Ain Shams University, Egypt.

**Zaki Taha** **zaki.taha@cis.asu.edu.eg**
Professor of Computer Science, Faculty of Computer and Information Sciences, Ain Shams University, Egypt.

**Walaa Gad** **walaagad@cis.asu.edu.eg**
Professor of InformationSystems**,** Faculty of Computer and Information Sciences, Ain Shams University, Egypt

**Abstract**

A survey is presented focused on using pose estimation techniques in Emotional recognition using various technologies normal cameras, and depth cameras for real-time, and the potential use of VR and inputs including images, videos, and 3-dimensional poses described in vector space. We discussed 19 research papers collected from selected journals and databases highlighting their methodology, classification algorithm, and the used datasets that relate to emotion recognition and pose estimation. A benchmark has been made according to their accuracy as it was the most common performance measurement metric used. We concluded that the multimodal Approaches overall made the best accuracy and then we mentioned futuristic concerns that can improve the development of this research topic.

## Introduction

Emotion recognition is one of the main vital tasks essential for having an intelligent system or application. Dealing with humans requires understanding their own emotions so that the human feels comfortable and the communication becomes more spontaneous which reflects on the efficiency of the service provided by the system/application. Emotions can be measured from multiple modalities like reading facial expressions, gesture detection, static posture, movement behavior, vocal tones, and text. When interacting with another human, you might know his current emotions from only seeing his face and sometimes the eyes can do the trick, or from his vocal tone, his posture - the way he is standing- or from the pattern of his movements, the gestures he is making or the context of his words whether those words are said or written - you can read an article and still visualize the emotions the writer has been through- or you can combine two or more modalities together which increases the efficiency of the human's predictions. Computer models are being trained to recognize the above models far above is the physical measurement which may include using sensors and actuators to measure physiological patterns that are hard for the computer to measure like measuring the heart rate, body temperature, and skin sensitivity(Picard, R.W. and Vyzas, E. and Healey, J., n.d.). Those extra modalities shall prepare the computer to be able to measure emotions accurately even more than humans, which is not currently reached. We will discuss the challenges being faced in this field and how some papers overcome those challenges. Some modalities can provide reliable measurements on their own or they may be used only to enhance the recognition of another modality and may not produce accurate results once used by themselves. In this paper, Our main focus will be on using the Pose estimation modality or posture recognition to measure the emotions of the human interacting with affective systems. The body posture or the pose can be detected from static images taken by a camera, image sequences (captured from videos) whether they are previously captured or provided in real-time, using a depth camera like Kinect which is usually used in providing real-time data, or finally using the Virtual reality technology which is usually real-time also. The images provide 2D coordinate system data unless a 2D to 3D conversion algorithm is implemented and that provides us with 3D coordinate system data or by using simply the depth camera or a VR device and sometimes it shall be equipped with external sensors to provide a full body detection including the lower body.

**Research Question:** What techniques and methodologies are used in literature to detect emotions through posture recognition?

**Objectives:**

- Observe how frequently each technology is Used.
- List the measurement metric of each methodology.
- explore the possibility of using Virtual Reality in the task of emotion recognition through posture detection.

Those keywords were chosen while doing the systematic review to be used in the academic databases and journals: Emotion Recognition/Detection AND Posture/Pose. The Virtual Reality keyword shall be used later in the paper classification step. The review shall be held from year 2019 to 2023. After the systematic review, we noticed the absence of Virtual reality usage and one of the main objectives of this survey paper was to explore the possibility of using Virtual reality technology in Pose detection so we added the Virtual Reality journal to the above, when those queries were used "Pose in Virtual Reality", "Pose estimation in Virtual Reality", "Pose detection in Virtual Reality" no results were found till 5/2024 but by combining the Pose and Virtual reality keywords we reached 184 research article which was refined for relevance according to their title and abstract.